\documentclass[sigconf]{acmart} 
\usepackage{booktabs} 
\usepackage{amsthm}

\newtheorem*{remark}{Remark}

\setcopyright{acmcopyright}

\usepackage{todonotes}
\usepackage{comment}
\specialcomment{auxproof}
{\mbox{}\newline\textbf{BEGIN: AUX-PROOF}\dotfill\newline}
{\mbox{}\newline\textbf{END: AUX-PROOF}\dotfill\newline}
\excludecomment{auxproof}

\usepackage{siunitx}

\copyrightyear{2024}
\acmYear{2024}
\setcopyright{rightsretained}
\acmConference[SAC '24]{The 39th ACM/SIGAPP Symposium on Applied Computing}{April 8--12, 2024}{Avila, Spain}
\acmBooktitle{The 39th ACM/SIGAPP Symposium on Applied Computing (SAC '24), April 8--12, 2024, Avila, Spain}\acmDOI{10.1145/3605098.3636014}
\acmISBN{979-8-4007-0243-3/24/04}

\begin{document}
\title{Temporal Logic Formalisation of ISO 34502 Critical Scenarios: Modular Construction with the RSS Safety Distance}
\titlenote{The work is partially supported by JST ERATO HASUO Metamathematics for Systems Design Project (No. JPMJER1603), JST START Grant (No. JPMJST2213), and JST CREST Grant (No. JPMJCR2012). J.\ Reimann is supported by a fellowship of the German Academic Exchange Service (DAAD).}
\subtitle{}
  
\renewcommand{\shorttitle}{Temporal Logic Formalisation of ISO 34502 Critical Scenarios}

\makeatletter

\newcommand\addauthornote[1]{%
  \if@ACM@anonymous\else
    \g@addto@macro\addresses{\@addauthornotemark{#1}}%
  \fi}

\newcommand\@addauthornotemark[1]{\let\@tmpcnta\c@footnote
   \setcounter{footnote}{#1}\addtocounter{footnote}{-1}
    \g@addto@macro\@currentauthors{\footnotemark\relax\let\c@footnote\@tmpcnta}}

\makeatother

\author{Jesse Reimann}
\authornote{Equal contribution}
\authornote{Work done during the stay at National Institute of Informatics (NII), Tokyo, Japan}
\orcid{...}
\affiliation{
  \institution{TU Delft}
  \city{Delft}
  \country{The Netherlands}
}

\author{Nico Mansion}
\addauthornote{1}
\addauthornote{2}
\orcid{...}
\affiliation{
  \institution{Saarland University}
  \city{Saarbr\"ucken}
  \country{Germany}
}

\author{James Haydon}
\affiliation{
 \institution{National Institute of Informatics}
  \city{Tokyo}
  \country{Japan}
}

\author{Benjamin Bray}
\affiliation{
  \institution{National Institute of Informatics}
  \city{Tokyo}
  \country{Japan}
}

\author{Agnishom Chattopadhyay}
\orcid{...}
\addauthornote{3}
\affiliation{
  \institution{Rice University}
  \city{Houston}
  \state{Texas}
  \country{USA}
}

\author{Sota Sato}
\orcid{...}
\affiliation{
  \institution{NII \& SOKENDAI}
  \city{Tokyo}
  \country{Japan}
}

\author{Masaki Waga}
\orcid{...}
\affiliation{
  \institution{Kyoto University \& NII}
  \city{Kyoto}
  \country{Japan}
}

\author{\'Etienne Andr\'e}
\orcid{...}
\affiliation{
  \institution{Université Sorbonne Paris Nord, \\ LIPN, CNRS, UMR 7030, F-93430 Villetaneuse, France}
  \country{}
}

\author{Ichiro Hasuo}
\orcid{0000-0002-8300-4650}
\affiliation{
  \institution{NII \& SOKENDAI}
  \city{Tokyo}
  \country{Japan}
}

\author{Naoki Ueda}
\orcid{...}
\affiliation{
  \institution{Mitsubishi Electric Corporation}
  \city{Kamakura}
  \country{Japan}
}

\author{Yosuke Yokoyama}
\orcid{...}
\affiliation{
  \institution{Mitsubishi Electric Corporation}
  \city{Kamakura}
  \country{Japan}
}

\renewcommand{\shortauthors}{J.\ Reimann, N.\ Mansion, et al.}

\begin{abstract}
As the development of autonomous vehicles progresses, efficient safety assurance methods become increasingly necessary. Safety assurance methods such as monitoring and scenario-based testing call for \emph{formalisation} of driving scenarios. In this paper, we develop a temporal-logic formalisation of 
an important class of critical scenarios in the ISO standard 34502. We use \emph{signal temporal logic (STL)} as a logical formalism. Our formalisation has two main features: 1) \emph{modular composition} of logical formulas for systematic and comprehensive formalisation (following the compositional methodology of ISO 34502); 2) use of the \emph{RSS distance} for defining danger. We find our formalisation comes with few parameters to tune thanks to the 
RSS distance. 
We experimentally evaluated our formalisation; using its results, we discuss the validity of our formalisation and its stability with respect to the choice of some parameter values.

\end{abstract}

%
%



\maketitle

\section{Introduction}

To increase social acceptance of automated driving vehicles (ADVs), addressing safety concerns is vital. The safety vision of the UNECE World Forum for Harmonization of Vehicle Regulations (WP.29) states that the level of safety should be such that ADVs ``shall not cause any traffic accidents resulting in injury or death that are reasonably foreseeable and preventable''~\cite{WP29}.

One path towards realising this vision is scenario-based testing of vehicle controllers~\cite{scenariobased_litreview}. For this approach, a library of critical scenarios is compiled, and the behaviour of the path-planning algorithm in these situations is observed through simulations. Compared to a posteriori analysis of recorded driving data, this allows for cost-efficient testing and adjustments of controllers, as it is not necessary to record new driving data after algorithmic changes. 

Efforts towards standardising test-scenario based ADV safety evaluations 
include ISO Standard 34502~\cite{ISO34502}. This standard derives relevant scenarios 
by systematically identifying risk factors related to vehicle perception, traffic conditions, and vehicle control. Combining these factors yields a large number of critical scenarios. 

Most efforts towards specifying critical scenarios use a combination of natural language descriptions and suitably chosen parameter value ranges, see e.g.\ \cite{cutout_scenario_generation}. An alternative is \emph{formalisation} of critical scenarios, in some formal language with a rigorously defined semantics, so that 1) the content of each scenario is mathematically well-defined, and 2) those scenarios can be \emph{mechanically processed} by software. A typical example of such processing is \emph{monitoring}, i.e.\ detecting occurrences of scenarios automatically in a driving log. 

A class of formal languages particularly suitable for this purpose is \textit{temporal logics}, 
pioneered by Pnueli~\cite{pnuelitemporallogic}. Temporal logics can be thought of as extensions of propositional logics with so-called \emph{temporal operators},
allowing for statements that refer to temporal relationships such as
 \textit{always} or \textit{eventually} in the future. Well-known examples of  temporal logics are LTL, CTL and CTL$^{\star}$; they all have discrete notions of time. For the purpose of formalising critical scenarios, their extensions to continuous time are suited, such as \emph{metric temporal logic (MTL)}~\cite{koymans_mtl_1990} and \emph{signal temporal logic (STL)}~\cite{stl_definition}.

In this paper, we provide a logical formalisation in STL of an important class of critical scenarios in ISO 34502, namely \emph{traffic disturbance scenarios} for general vehicles on highways. Our formalisation has two unique characteristics. The first one is its modularity:  a common template of the formulas for different scenarios is first fixed, and then the component formulas in the template are instantiated. This construction---it mirrors the top-down compositional methodology 
in ISO 34502---aids systematic, comprehensive formalisation.  The second characteristic is the use of the \emph{RSS distance}~\cite{Shalev-Shwartz_Shammah_Shashua_2018}, a distance considered safe between two moving vehicles, for defining danger. We find that this makes our formalisation more stable by reducing the number of parameters to be tuned. We also report our use of our tool called \emph{STL Debugger} to improve our workflow.

In general, a formalisation of critical scenarios in temporal logic has two main usages: \emph{scenario-based testing} (where a vehicle controller and a scenario are given and we search for a traffic situation in which the controller behaves as specified in the scenario) and \emph{monitoring} (where a driving log and a scenario are given and we search for those segments of the log which match the scenario). We note that there are a body of sophisticated tools  that accept STL formulas, for scenario-based testing~\cite{DBLP:conf/cav/ZhangHA19,DBLP:journals/tcad/ZhangESAH18,Donze10,AnnpureddyLFS11} and for monitoring~\cite{DBLP:phd/hal/Ulus18}.  Our formalisation can be readily utilised by these tools.

\paragraph{Contributions} Our technical contributions are as follows. 
\begin{itemize}
    \item We provide a set of logical formulas describing traffic disturbance scenarios based on ISO 34502.
    \item The two unique characteristics of our formalisation (namely modularity and the use of the RSS distance) suggest a general methodology for formalising various driving scenarios, beyond those in ISO 34502.
    \item  We evaluate the adequacy of the traffic disturbance scenario descriptions in ISO 34502 and propose an extension.
    \item Experimental evaluation of our formulas demonstrates that we detect nearly all traffic disturbances in a given dataset.
\end{itemize}

\paragraph{Organisation}
We introduce the conceptual and logical framework of our formalisation in \S\ref{sect: Preliminaries} and \S\ref{sect: traffic scenario formalisation}. Our formalisations \textit{ISO34502-STL} and \textit{ISO34502-STL-ext} and the formulas used to construct them are introduced in \S\ref{sect: subformulas}. 
 We describe the experimental evaluation of our formulas in \S\ref{sect: experiments}, and summarise our results in \S\ref{sect: conclusion}.

Towards of our formalisation, we used our tool 
 STL Debugger for deriving and debugging STL formulas. The tool and our use of it is discussed in
Appendix~\ref{sect: debugger}.

\paragraph{Related work}
The RSS framework~\cite{Shalev-Shwartz_Shammah_Shashua_2018} is based on the assumption that all vehicles comply with pre-defined proper responses to potentially critical scenarios. In our work, we use the RSS framework to obtain a safety metric suitable to the logical formalisation of critical scenarios without the notion of responsibility implied by the proper responses. Some proper responses 
from~\cite{Shalev-Shwartz_Shammah_Shashua_2018} were formalised in STL in~\cite{formalised_RSS}. Suitable parameter ranges for calibrating the RSS distance are studied e.g.\ in~\cite{KonigshofOSS22}. 
On the use of RSS for \emph{verifying} safety, recent developments are found e.g.\ in~\cite{ga_rss}. 

There are several efforts to formalise traffic rules in temporal logic, such as selected laws from the German StVO relating to interstate roads~\cite{interstate_althoff}, dual carriageways~\cite{knoll_machine_interpretability}, and intersections~\cite{intersection_althoff}, as well as marine laws~\cite{marine_althoff}. These works formalise \emph{rules} to be obeyed and thus use the \textit{always} modality often, while ours is about \emph{disturbance} and thus primarily uses the \emph{eventually} and \emph{until} modalities.

There are existing methods of detecting critical scenarios that rely on metrics such as time-to-collision (TTC) or jerk. In~\cite{streetwise,highDMonitoring,highDsakuracomparison}, critical scenarios resulting from cut-ins and braking are detected in monitored data. A more general workflow based on detecting critical thresholds for jerk and time-to-collision is proposed in~\cite{song_critical_2023}. The Virtual Assessment of Automation in Field Operation (VAAFO)~\cite{vaafo} method uses, among other metrics, a weighted sum across potential driving paths is used to identify critical scenarios where a human driver is assisted by a Highly Automated Driving (HAD) system. Our logical formalisation is independent of our specific choice of a safety metric (namely the RSS distance); use of other metrics such as TTC is well possible. That said, we believe the choice of  RSS distance is another contribution of ours: it is known to be one of the advanced and robust metrics, and can easily be accommodated in temporal-logic formalisation.

\section{Preliminaries}\label{sect: Preliminaries}
\subsection{ISO Standard 34502}\label{subsect: prelim ISO}ISO 34502~\cite{ISO34502} is a proposal of safety evaluation procedures of automated driving systems, based on a joint industry effort by Japanese automotive manufacturers~\cite{jamareport}. \textit{Critical scenarios}, defined in ISO 34502 as ``scenario[s] including one or more risk factors'', 
are systematically constructed in a top-down manner by combining three categories of risk factors. They correspond to the three phases in the common automated driving pipeline, namely perception, planning, and control: 

A \emph{perception disturbance} includes intrinsic or extrinsic factors that can cause disturbance of sensors and cameras, such as light reflections or roadside objects. \emph{Traffic disturbances} arise from road geometry or the behaviour of other traffic participants, such as dangerous cut-ins or 
braking. \emph{Vehicle control disturbances} describe internal or external factors that impact the feasibility of certain 
maneuvers, such as vehicle weight distribution 
or strong winds.

By refining and systematically combining these risk factors, a large number of critical scenarios arises. This can be considered a set of reasonably foreseeable critical scenarios in which an ADV is expected to avoid collisions or minimise unavoidable accidents.

In this paper, we focus on \textit{traffic disturbance scenarios} as defined in ISO 34502 for general vehicles on highways. They are composed of the following three components:

The first component is \textit{road geometry}: On highways, one may assume that there are no intersections, roundabouts, or unstructured roads. The relevant road shapes for highways are thus non-intersection segments (straight or curved), merge zones where two roads merge into a single road, and departure zones, where an additional road branches off the main road.

The second component is \textit{subject vehicle (SV) behaviour}. It is classified into either keeping the current lane (\textit{lane keep}) or changing lanes (\textit{lane change}). It is assumed that all vehicles behave reasonably, hence backwards driving or U-turns are not considered.

\begin{figure}
    \includegraphics[width=0.45\textwidth]{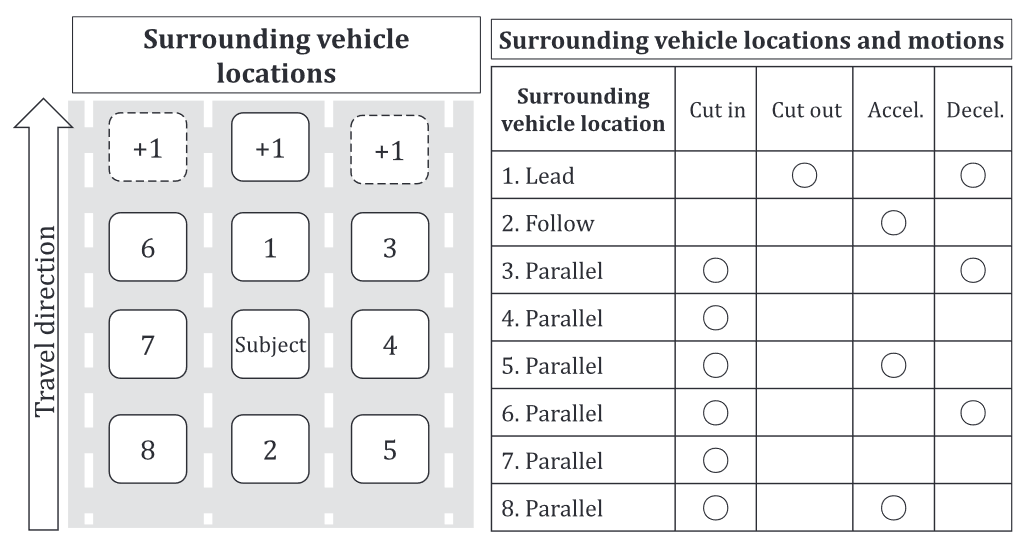}
    \caption{Left: Numbering of possible $\mathit{POV}$ locations relative to $\mathit{SV}$. The fields numbered as "+1" are only relevant in three-vehicle scenarios. Right: Combinations of $\mathit{POV}$ positions and behaviours that may cause critical scenarios. These are from~\cite{ISO34502}.}\label{fig:vehiclepositions}
\end{figure}

The last of the three components is \textit{principle other vehicle (POV) behaviour}: The behaviour of the vehicles surrounding $\mathit{SV}$ is assumed to cause a dangerous situation for the vehicles involved. Here, these behaviours are classified into either 1) changing into the lane of $\mathit{SV}$ (\textit{dangerous cut-in}), 2) leaving the lane of $\mathit{SV}$ (\textit{dangerous cut-out}), 3) driving dangerously fast behind $\mathit{SV}$ (\textit{acceleration}), or 4) driving dangerously slow in front of $\mathit{SV}$ (\textit{deceleration}). The location of $\mathit{POV}$ relative to $\mathit{SV}$ is used to exclude non-critical behaviours, e.g. $\mathit{SV}$ will not be endangered if $\mathit{POV}$ drives faster in front of it. For critical combinations of $\mathit{POV}$ behaviours and vehicle locations, see Fig.~\ref{fig:vehiclepositions}.

Combining these three components results in 24 traffic disturbance scenarios, see Table~\ref{table: scenarios}. While $\mathit{SV}$ may be surrounded by several $\mathit{POV}$s, traffic disturbance scenarios can generally be composed of subscenarios involving only two vehicles, thus a formalisation of the 24 scenarios in Table~\ref{table: scenarios} is the foundation of more comprehensive sets of formalised scenarios, including many-vehicle scenarios. 

However, a \textit{three-vehicle cut-out} is included in Table~\ref{table: scenarios}, namely Scenario~2: There are two vehicles ahead of $\mathit{SV}$ in the same lane as $\mathit{SV}$. The $\mathit{POV}$ directly ahead of $\mathit{SV}$ changes lanes, whereas the other $\mathit{POV}$ remains in the lane. This scenario is critical when the $POV$ remaining in the lane is slower than $\mathit{SV}$ could reasonably expect, causing risk of collision. 

\begin{table}
    \caption{General vehicle traffic disturbance scenarios on highways in ISO Standard 32502~\cite{ISO34502} (the table is from~\cite{ISO34502}). The white vehicle represents $\mathit{SV}$, the darker vehicles the $\mathit{POV}$s, the arrows indicate the (intended) motion of the vehicles. The rightmost four columns refer to $\mathit{POV}$ behaviour.}
    \label{table: scenarios}
    \includegraphics[width=0.45\textwidth]{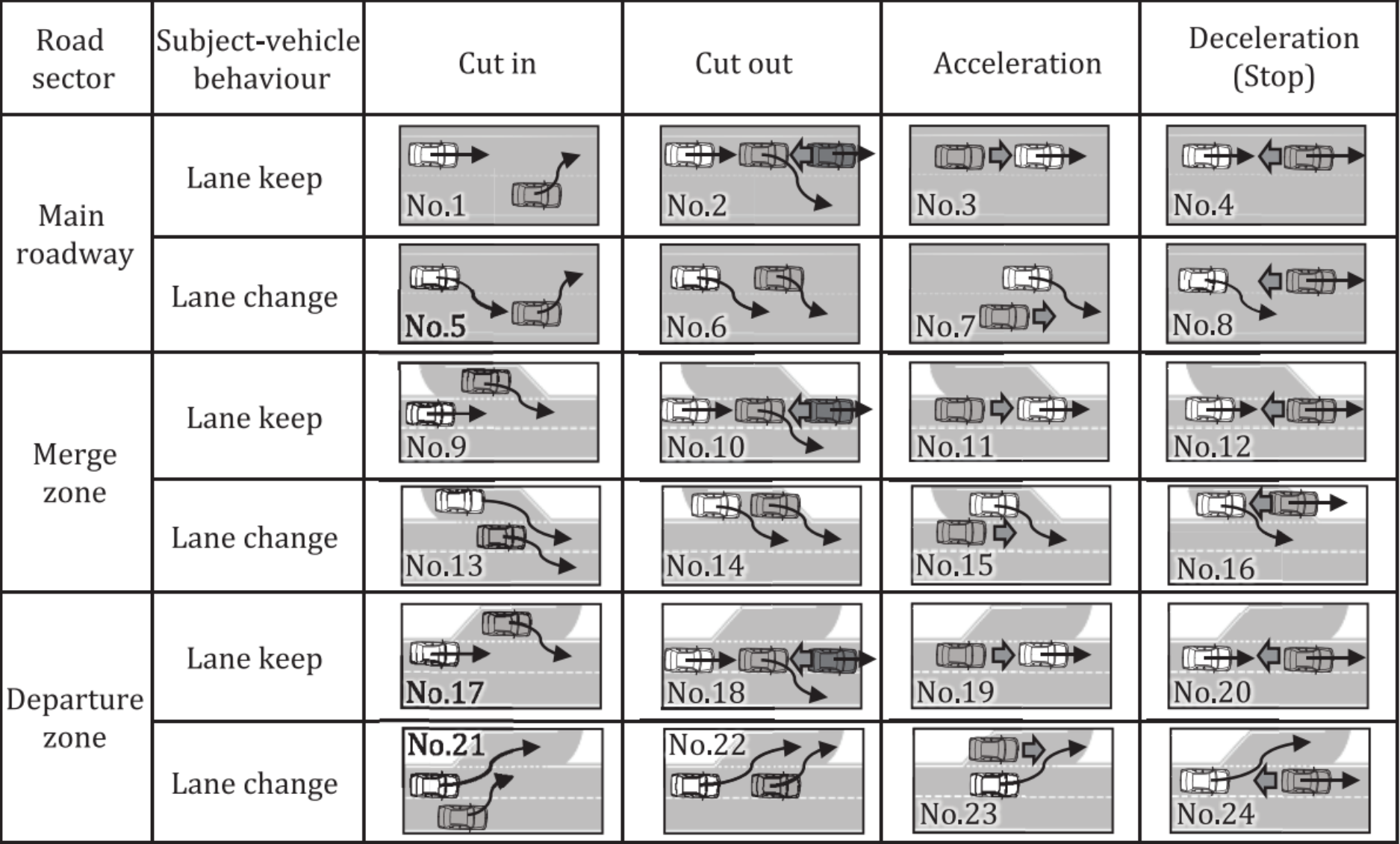}
\end{table}

\subsection{STL}\label{subsect:STL}

We formalise the traffic disturbance scenarios in ISO 34502 using \emph{signal temporal logic} (STL) in a rigorous manner that can be used for monitoring purposes and scenario generation. 
STL~\cite{stl_definition} is an extension of temporal logics such as LTL (see e.g.~\cite{BaierK08}) that is designed to specify the behaviour of continuous-time signals in a succinct manner that nevertheless maintains readability for humans. The familiar propositional logic using the operators $\neg,\vee,\wedge$ is extended by adding temporal operators $\mathcal{G},\mathcal{F},\mathcal{U}$ denoting the temporal notions \textit{always} (or \emph{globally}), \textit{eventually} (or \emph{finally}), and \textit{until}. 
 \emph{Formulas} in STL are constructed inductively from the following grammar, where we let $f: \mathbb{R}^n \to \mathbb{R}$, $x \in \mathbb{R}^n, n\in \mathbb{N}$, and $J \subseteq \mathbb{R}$ an interval. 
\begin{align*}
    \varphi &::= \top \mid \bot \mid f(x) > 0 \mid \neg \varphi \mid \varphi_1 \vee \varphi_2 \mid  \\
    &\quad \varphi_1 \wedge \varphi_2 \mid \varphi_1 \mathcal{U}_J \varphi_2 \mid \mathcal{G}_J \varphi \mid \mathcal{F}_J \varphi.
\end{align*}
Here, the subscript $J$ denotes the restriction of an operator to a time interval $J$ relative to the current time. As an example, consider a formula $\mathcal{G}_{[2,3]}(v > 5)$. This formula is true at time $T$ if the value of $v$ is greater than $5$ always from time $T+2$ to time $T+3$. If no interval $J$ is specified, the formula is treated as if $J=[0,\infty)$.

The meaning of STL formulas is 
defined by their \emph{Boolean semantics}, inductively defined in Table~\ref{table: booleansemantics}. The relationship $\sigma\models\varphi$ means that the formula $\varphi$ is \emph{true} under a signal ${\sigma: [0, \infty) \to \mathbb{R}^n}$, $n\in\mathbb{N}$. 

Particularly important in our formalisation are the semantics of the \textit{until} operator $\mathcal{U}$. For a formula $\varphi_1\mathbin{\mathcal{U}}\varphi_2$ to be true at time~$T$, it must hold that 1) $\varphi_2$ is true at some time $T'\ge T$ and 2) $\varphi_1$ is constantly true throughout the time interval $[T,T')$. Note that $\varphi_1$ is not required to be true at the time $T'$ when $\varphi_2$ first becomes true; the formula $\varphi_1\mathbin{\mathcal{U}}\varphi_2$ is thus trivially true at time $T$ if $\varphi_2$ is true at time $T$. This becomes important in our formalisation, see \S\ref{subsec:danger}.

\begin{table}[tbp]
    \caption{\label{table: booleansemantics}Boolean semantics of STL}
    \begin{tabular}{lll}
    $\sigma \vDash \top$ & $\leftrightarrow$ & \textit{true} \\
    $\sigma \vDash \bot$ & $\leftrightarrow$ & \textit{false} \\
    $\sigma \vDash (f(x) >0)$ & $\leftrightarrow$ & $f(\sigma(0)) > 0$ \\
    $\sigma \vDash \neg\varphi$ & $\leftrightarrow$ & $\sigma\nvDash\varphi$ \\
    $\sigma \vDash \varphi_1 \vee \varphi_2$ & $\leftrightarrow$ & $\sigma\vDash\varphi_1$ or $\sigma\vDash\varphi_2$ \\
    $\sigma \vDash \varphi_1 \wedge \varphi_2$ & $\leftrightarrow$ & $\sigma\vDash\varphi_1$ and $\sigma\vDash\varphi_2$ \\
    $\sigma \vDash \varphi_1 \mathcal{U}_J \varphi_2$ & $\leftrightarrow$ & $\exists t\in J:$ $\sigma(\cdot+t)\vDash\varphi_2$ \\
    & & $\quad$ and $\sigma(\cdot+t')\vDash\varphi_1$ for all $t'<t$ in $J$ \\
    $\sigma \vDash \mathcal{F}_J \varphi$ & $\leftrightarrow$ & $\exists t\in J:$ $\sigma(\cdot+t)\vDash\varphi$ \\
    $\sigma \vDash \mathcal{G}_J \varphi$ & $\leftrightarrow$ & $\forall t\in J:$ $\sigma(\cdot+t)\vDash\varphi$ \\
    \end{tabular}
    \end{table}

\subsection{Road Network and Vehicle Configuration}\label{subsect:vehicleConf}
To model our road network, we use lanelets~\cite{lanelets} and their logical formalisation in~\cite{interstate_althoff}, so that our formulas can be applied to a wide range of road shapes. 
A rigorous description of our road network can be found in Appendix~\ref{appendix:roadNetwork}. 

\begin{figure}
    \includegraphics[width=0.35\textwidth]{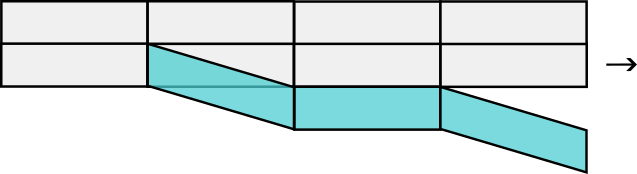}
    \caption{A sample road section (departure zone) constructed from lanelets forming two main road lanes (grey, $\mathit{attr}(l)=\mathit{main}$) and one departure lane (blue, $\mathit{attr}(l)=\mathit{departure}$). All depicted lanelets satisfy $\mathit{zone}(l)=\mathit{ departZone}$. The driving direction is indicated by the arrow. The departure lane is adjacent to the lower main road lane but not adjacent to the upper main road lane.}
    \label{fig:lanelets}
\end{figure}

In the current work, the relevance of lanelets is quite limited: they are used only in the very beginning of our workflow, for the purpose of mapping physical positions to lane coordinates. After that, the description of a disturbance scenario is  independent of road geometry; this is seen e.g.\ in Table~\ref{table: scenario_formulas}.
In any case, 
here we give an informal overview of lanelets.
For an illustration, see Fig.~\ref{fig:lanelets}.

Lanelets are atomic road segments described by their (piecewise straight) left and right bounds. Each lanelet has an associated \emph{attribute}, denoting whether it belongs to a main road lane or to a merge/departure lane, and an associated \emph{zone}, stating whether the lanelet is part of an overall merge/departure zone of the road. 

A \emph{lane} is then defined as the set of all lanelets of the same attribute that pre- or succeed each other. Note that by this construction, a single lane cannot branch out and lead in two different directions; however, two lanes may overlap. We call two lanes $L_1$ and $L_2$ \emph{adjacent} if there are some lanelets $l_1\in L_1$, $l_2\in L_2$, such that $l_1$ and $l_2$ share a boundary line but do not overlap.

\begin{figure}
    \includegraphics[width=0.25\textwidth]{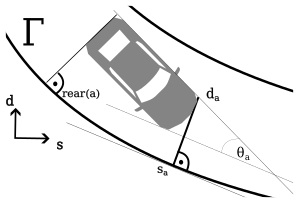}
    \caption{Curvilinear coordinates, with the reference path $\Gamma$ being the lower road boundary. Here, $s_a$ denotes the coordinate of the vehicle $a$ along the $\Gamma$, $d_a$ the distance of $a$ from $\Gamma$, and $\theta_a$ the orientation of the vehicle relative to $\Gamma$. See Section~\ref{subsect:vehicleConf} for the definition of $\mathit{rear}(a)$.}
    \label{fig:curvilinear}
\end{figure}

We further follow~\cite{interstate_althoff} and model the dynamics of our vehicles as point masses described by $x=(s,v,a,d,\theta)\in\mathbb{R}^5$. Here $s$ denotes the position of the vehicle along a fixed reference path $\Gamma$ (typically the boundary of an outer lane), $v$ its velocity, and $a$ its acceleration. The lateral distance of the vehicle to 
 $\Gamma$ is denoted by $d$, and its orientation relative to $\Gamma$ by $\theta$, see Fig.~\ref{fig:curvilinear}. 

For different vehicles, we use subscripts for the variables: $x_{\square}=(s_{\square},v_{\square},a_{\square},d_{\square},\theta_{\square})$ for a vehicle $\square$, where $\square$ is typically either $\mathit{SV}$ or $\mathit{POV}$; e.g.\ $v_{\mathit{SV}}$ is the velocity of $\mathit{SV}$. In formulas, we refer to $x_{\mathit{SV}}$ and $x_{\mathit{POV}}$ simply as $\mathit{SV}$ and $\mathit{POV}$. When no confusion with the acceleration is possible, we refer to generic vehicles as $a$ and $b$.

To define the occupancy of a vehicle, we associate a box of length $\mathit{length}(a)$ and width $\mathit{width}(a)$ with each vehicle $a$. We choose the front-left corner of this box as the point of which the location and dynamics are tracked, and define $\mathit{occ}(a,l)$ to be true if any part of the box associated with vehicle $a$ is within the road segment described by lanelet $l$, and false otherwise. Furthermore, we can define the coordinates $\mathit{front}(a)$, $\mathit{rear}(a)$ as the  projection of the front and rear end of the vehicle onto the reference path. By construction, we have $\mathit{front}(a)=s_a$. On highways, we may assume that road segments are locally straight, hence we assume $\mathit{rear}(a)\approx s_a-\mathit{length}(a)$.

Longitudinal and lateral velocities of the vehicle relative to the reference path are given by $v_{long}\;:=\;v\sin(\theta)$, $v_{\mathit{lat}}\;:=\;v\cos(\theta)$.

\section{Traffic Disturbance Scenario Formalisation}\label{sect: traffic scenario formalisation} 
\begin{table*}[]
    \caption{Formalisation \textit{ISO34502-STL} of general vehicle traffic disturbance scenarios of ISO 34502 (cf. \S\ref{subsect: prelim ISO}) in STL. For definitions of \textit{initSafe} and \textit{danger} see \S\ref{subsec:danger}, for all other subformula definitions see the section indicated in the corresponding column heading. For a discussion of the formulas see \S\ref{subsec: ISO34502-STL}.}
    \label{table: scenario_formulas}
    \small
    \begin{tabular}{ll|clll}
    \multicolumn{6}{l}{$\mathit{scenario}_{i}(\mathit{SV},\mathit{POV},L) \;:=\; \mathit{initSafe}(\mathit{SV},\mathit{POV}) \wedge \mathit{roadSector}_i(\mathit{SV},\mathit{POV})\wedge \mathit{disturb}_i(\mathit{SV},\mathit{POV},L)$, $i=1,\dotsc,24$ (cf.\ this is (\ref{scenarioAsInitSafePlusDisturb}). $\mathit{initSafe}$ is from \S\ref{subsec:danger})} \\
    \multicolumn{6}{l}{$\mathit{disturb}_i(\mathit{SV},\mathit{POV},L) \;:=\; \mathit{initialCondition}_i(\mathit{SV},\mathit{POV},L) \wedge \mathit{behaviourSV}_i(\mathit{SV},\mathit{POV},L)\wedge \mathit{behaviourPOV}_i(\mathit{POV},\mathit{SV},L)$ , $i=1,\dotsc,24$ (cf. (\ref{disturbFormat}) in \S\ref{sect: traffic scenario formalisation})} \\
    \toprule
     $i$ & $\mathit{roadSector}_i$ (cf. \S\ref{subsec: vehiclepos}) & $i$ &$\mathit{initialCondition}_i$ (cf.\S\ref{subsec: relbtwnvehicles}) & $\mathit{behaviourSV}_i$ (cf.\S\ref{sect: svbehaviour}) & $\mathit{behaviourPOV}_i$ (cf.\S\ref{sect: povbehaviour}) \\
    \toprule
     && 1& $\top$    &   $\mathit{laneKeep}(\mathit{SV},L)$    &  $\mathit{cutIn}(\mathit{POV},\mathit{SV})$    \\
     &  & &  &   $\quad\mathbin{\mathcal{U}} \mathit{danger}(\mathit{SV},\mathit{POV})$    & \\
     && 2&$\mathit{sameLane}_3(\mathit{SV},\mathit{POV}_1,\mathit{POV}_2,L)$      &  $\mathit{laneKeep}(\mathit{SV},L)  $  &   $\mathit{leavingLane}(\mathit{POV}_1,L) $   \\
     && &$\quad\wedge \mathit{aheadOf}(\mathit{SV},\mathit{POV}_1) $ & $\mathbin{\mathcal{U}} (\neg \mathit{sameLane}(\mathit{SV},\mathit{POV}_1,L))$ & $\quad\wedge(\mathit{laneKeep}(\mathit{POV}_2,L) $ \\
     & && $\quad\wedge \mathit{aheadOf}(\mathit{POV}_1,\mathit{POV}_2)$ &&$\quad\quad \mathbin{\mathcal{U}} (\neg \mathit{sameLane}(\mathit{POV}_2,\mathit{POV}_1,L)$   \\
     &&&&& $\quad\quad\quad\wedge \mathit{danger}(\mathit{SV},\mathit{POV}_2)))$ \\
     & &  3&$\mathit{aheadOf}(\mathit{POV},\mathit{SV})$ &   $\mathit{laneKeep}(\mathit{SV},L) $ & $\mathit{accel}(\mathit{POV},\mathit{SV},L) \mathbin{\mathcal{U}} \mathit{danger}(\mathit{SV},\mathit{POV})$ \\
     & &&$\quad \wedge (\mathit{sameLane}(\mathit{SV},\mathit{POV},L)$&$\quad\mathbin{\mathcal{U}} \mathit{danger}(\mathit{SV},\mathit{POV})$& \\
     & && $\quad\quad \vee \mathit{inAdjLanes}(\mathit{SV},\mathit{POV},L))$&& \\
     $1-8$ & $\mathit{mainRoad}(\mathit{SV},\mathit{POV})$ &  4&$\mathit{aheadOf}(\mathit{SV},\mathit{POV})$ &   $\mathit{laneKeep}(\mathit{SV},L)$ & $\mathit{decel}(\mathit{POV},\mathit{SV},L) \mathbin{\mathcal{U}} \mathit{danger}(\mathit{SV},\mathit{POV})$ \\
     & &&$\quad \wedge (\mathit{sameLane}(\mathit{SV},\mathit{POV},L)$&$\quad\mathbin{\mathcal{U}} \mathit{danger}(\mathit{SV},\mathit{POV})$& \\
     & &&$\quad\quad \vee \mathit{inAdjLanes}(\mathit{SV},\mathit{POV},L))$&& \\
     & & 5&$\top$ & $\mathit{leavingLane}(\mathit{SV},L)$ & $\mathit{cutIn}(\mathit{POV},\mathit{SV})$ \\
     &  &6& $\top$ & $\mathit{leavingLane}(\mathit{SV},L)$ & $\mathit{cutOut}(\mathit{POV},\mathit{SV},L)$\\
     & &7& $\mathit{aheadOf}(\mathit{POV}, \mathit{SV})$ & $\mathit{enteringLane}(\mathit{SV},L)$ & $\mathit{accel}(\mathit{POV},\mathit{SV},L) \mathbin{\mathcal{U}} \mathit{danger}(\mathit{SV},\mathit{POV})$ \\
     & & 8&$\mathit{sameLane}(\mathit{SV},\mathit{POV},L)$ &$\mathit{leavingLane}(\mathit{SV},L)$ & $\mathit{decel}(\mathit{POV},\mathit{SV},L) \mathbin{\mathcal{U}} \mathit{danger}(\mathit{SV},\mathit{POV})$\\
     &&&$\quad\wedge \mathit{aheadOf}(\mathit{SV},\mathit{POV})$&& \\
     \midrule
     $9$--$16$& $\mathit{mergeZone}(\mathit{SV},\mathit{POV})$&$9$--$16$& $\mathit{initialCondition}_{i-8}$ & $\mathit{behaviourSV}_{i-8}$ & $\mathit{behaviourPOV}_{i-8}$ \\
     \midrule
     $17$--$24$& $\mathit{departZone}(\mathit{SV},\mathit{POV})$&$17$--$24$& $\mathit{initialCondition}_{i-16}$ & $\mathit{behaviourSV}_{i-16}$ & $\mathit{behaviourPOV}_{i-16}$ \\
     \bottomrule
    \end{tabular}
\end{table*}

We follow the modular approach of ISO 34502 (\S\ref{subsect: prelim ISO}) in our STL formalisation of its traffic disturbance scenarios for general vehicles on highways (Table~\ref{table: scenarios}). That is, we first fix a \emph{template} of STL formulas that express traffic disturbance scenarios, and by varying the ``parameters'' (i.e.\ component subformulas) of the template, we systematically obtain the formalisation of different scenarios.

Our top-level formulas are $\mathit{scenario}_i$, for $i=1, 2, \dotsc, 24$, each of which represents one of the traffic disturbance scenarios in Table~\ref{table: scenarios}. The format for these formulas $\mathit{scenario}_i$ is the following:
\begin{align}\label{scenarioAsInitSafePlusDisturb}
\begin{array}{rl}
    scenario_i(\mathit{SV},\mathit{POV},L) \;:=\; &\mathit{initSafe}(\mathit{SV},\mathit{POV}) \\ &\wedge \mathit{roadSector}_i(\mathit{SV},\mathit{POV})\\ & \wedge \mathit{disturb}_i(\mathit{SV},\mathit{POV},L).
\end{array}
\end{align}
Here $\mathit{initSafe}(\mathit{SV},\mathit{POV})$ is a formula---common to all scenarios---stating that there is no hazard at the start of the scenario for the duration of a pre-defined time interval; its definition is given in~\S\ref{subsec:danger}. The formula $\mathit{disturb}_i(\mathit{SV},\mathit{POV},L)$ expresses the content of the actual disturbance and is specific to each traffic disturbance scenario. Here, $L$ denotes a lane relative to which the scenario is formalised.

We emphasise the importance of the formula $\mathit{initSafe}(\mathit{SV},\mathit{POV})$. In fact, it was absent in our early trials, causing most disturbance formulas to be trivially true for certain traces. In particular, this problem occurs with traces that either start with danger between the observed vehicles or in which the monitored interval leading up to danger is too short; we return to this discussion in \S\ref{subsec: experimentdiscussion}. 

The formula $\mathit{roadSector}_i(\mathit{SV},\mathit{POV})$ corresponds to the \emph{road geometry} component in \S\ref{subsect: prelim ISO},  i.e.\ main road, merge zone, or departure zone. For details, see \S\ref{subsec: ISO34502-STL}.

We label the set $\{\mathit{scenario}_i\mid i=1,\dotsc,24\}$ \textit{ISO34502-STL}.

The general structure of our disturbance formulas $\mathit{disturb}_i$ is 
\begin{align}\label{disturbFormat}
\begin{split}
\mathit{disturb}_i(\mathit{SV},\mathit{POV},L) &\;:=\; \mathit{initialCondition}_i(\mathit{SV},\mathit{POV},L) \\
 &\quad \wedge \mathit{behaviourSV}_i(\mathit{SV},\mathit{POV},L) \\
 &\quad \wedge \mathit{behaviourPOV}_i(\mathit{POV},\mathit{SV},L)
\end{split}
\end{align} 
corresponding roughly to the latter two of the three components in \S\ref{subsect: prelim ISO}, namely $\mathit{SV}$ behaviour and $\mathit{POV}$ behaviour.
Each component of the template (\ref{disturbFormat}) is as follows.
\begin{itemize}
 \item The formula $\mathit{initialCondition}_i(\mathit{SV},\mathit{POV},L)$ specifies initial conditions, e.g.\ initial lanes and vehicle order (see Fig.~\ref{fig:vehiclepositions}). In some traffic disturbance scenarios (such as in Scenarios $1$, $5$, or~$6$), no such initial conditions are necessary, in which case we set $\mathit{initialCondition}_i(\mathit{SV},\mathit{POV},L)=\top$. By the semantics of $\top$ (see Table~\ref{table: booleansemantics}), this subformula is thus trivially true.
 \item The formula $\mathit{behaviourSV}_i(\mathit{SV},\mathit{POV},L)$ corresponds to the \emph{SV behaviour} component in \S\ref{subsect: prelim ISO}, i.e.\ keeping the current lane $L$, or changing from lane $L$ into a different lane.
 \item The formula $\mathit{behaviourPOV}_i(\mathit{POV},\mathit{SV},L)$  corresponds to the \emph{POV behaviour} component in \S\ref{subsect: prelim ISO}, namely cut-in, cut-out, acceleration, and deceleration. Note that $\mathit{SV}$'s behaviour is also relevant here; the formula thus has $\mathit{SV}$ as its argument as well as the lane $L$ relative to which the behaviour of $\mathit{SV}$ is described. Notably, $\mathit{POV}$ behaviour formulas always contain a subformula of the form \begin{align*}
& \varphi_1(\mathit{SV},\mathit{POV}) \mathbin{\mathcal{U}} (\varphi_2(\mathit{SV},\mathit{POV}) \wedge \mathit{danger}(\mathit{SV},\mathit{POV}))
\\
&\text{or}\quad\mathcal{F}(\varphi_2(\mathit{SV},\mathit{POV}) \wedge \mathit{danger}(\mathit{SV},\mathit{POV})),
\end{align*} expressing that $\mathit{POV}$ behaviour is dangerous relative to $\mathit{SV}$.
\end{itemize}

The overall structure of the formula $\mathit{disturb}_i$ (\ref{disturbFormat}) is as follows. It requires that a general notion of danger is true at a certain stage (recall that $\varphi_{1}\mathbin{\mathcal{U}} \varphi_{2}$ requires $\varphi_{2}$ becoming true at some time); the formulas $\mathit{behaviourSV}_i$ and $\mathit{behaviourPOV}_i$ specify the behaviours of the vehicles in the period leading to that dangerous moment. Some additional initial conditions are expressed in $\mathit{initialCondition}_i$. 

There is, however, an initial condition that is common to all critical scenarios, namely that danger must happen after no danger has been observed for some time. This is entirely and exclusively described by $\mathit{initSafe}$ in (\ref{scenarioAsInitSafePlusDisturb}).

The STL formulas $\mathit{disturb}_i$ for the traffic disturbance scenarios in Table~\ref{table: scenarios} are collected in Table~\ref{table: scenario_formulas}. The subformulas occurring there (such as e.g.\ \textit{laneKeep} and \textit{cutIn}) are introduced in \S\ref{sect: subformulas}; after that we will review the formulas and see that they indeed correspond to the natural language descriptions in the ISO 34502 standard.

In the following section, we describe the formulas from which our formalisation of the traffic disturbance scenarios is composed. 
\section{Definitions of Formulas}\label{sect: subformulas}

\subsection{Vehicle Positions}\label{subsec: vehiclepos}
We say that a vehicle $a$ is in a given lane $L$ if it occupies any lanelet of that lane, 
where $\mathit{occ}(a,l)$ is from \S\ref{subsect:vehicleConf}.
 \begin{equation*}
    atLane(a,L)\;:=\;\textstyle\bigvee_{l\in L}(\mathit{occ}(a,l)).
\end{equation*}
A vehicle $a$ may occupy multiple lanes at once, for example during a lane change.
To characterise the road sector a vehicle $a$ is in, we refer to the zone of the lanelets the vehicle occupies, where the disjunction is indexed over all lanelets $l$ with the corresponding associated zone:
\begin{align*}
\begin{array}{rl}
     \mathit{onMainRoad}(a)&\;:=\; \textstyle\bigvee_{\mathit{zone}(l)=\mathit{mainZone}}  \mathit{occ}(a,l), \\
    \mathit{inMergeZone}(a)&\;:=\;\textstyle\bigvee_{\mathit{zone}(l)=\mathit{mergeZone}}  \mathit{occ}(a,l), \\
    \mathit{inDepartZone}(a)&\;:=\;\textstyle\bigvee_{\mathit{zone}(l)=\mathit{ departZone}}  \mathit{occ}(a,l). 
\end{array}
\end{align*}
We can now define the relevant road sector for a scenario by
\begin{align*}\small
\begin{array}{rl}
     \mathit{mainRoad}(\mathit{SV},\mathit{POV})&:=\; 
        \mathit{onMainRoad}(\mathit{SV})\wedge \mathit{onMainRoad}(\mathit{POV}), \\
        \mathit{mergeZone}(\mathit{SV},\mathit{POV})&:=\; 
        \mathit{inMergeZone}(\mathit{SV})\vee \mathit{inMergeZone}(\mathit{POV}), \\
        \mathit{ departZone}(\mathit{SV},\mathit{POV})&:=\; {\mathit{inDepartZone}(\mathit{SV})\vee \mathit{inDepartZone}(\mathit{POV}).}
\end{array}
\end{align*}

\subsection{Relationship between Vehicles}\label{subsec: relbtwnvehicles}
We express a difference in (longitudinal) position or velocity of two vehicles $a$ and $b$ by
\begin{align}
    \mathit{aheadOf}(a,b)&\;:=\;\mathit{front}(a) \le \mathit{rear}(b),
     \label{eq:aheadOfOriginal}\\
    \mathit{fasterThan}(a,b)&\;:=\; v_a < v_b, \nonumber
\end{align}
using the vehicle parameters \textit{front}, \textit{rear}, and $v$ defined in \S\ref{subsect:vehicleConf}. Using \textit{atLane} from \S\ref{subsec: vehiclepos} and $\mathit{adjLanes}(L)$ from Appendix~\ref{appendix:roadNetwork}, we further define formulas expressing that vehicles $a,b,c$ are in the same lane or in adjacent lanes as
\begin{align*}
    &\mathit{sameLane}(a,b,L) := atLane(a,L)\wedge atLane(b,L),\\
    &\mathit{inAdjLanes}(a,b,L) := \mathit{atLane}(a,L) \wedge \textstyle\bigvee_{L'\in\mathit{adjLanes}(L)}\mathit{atLane}(b,L'),\\
    &\mathit{sameLane}_3(a,b,c,L) := sameLane(a,b,L)\wedge sameLane(b,c,L).
\end{align*}

\subsection{Danger} 
\label{subsec:danger}
In Table~\ref{table: scenario_formulas}, 
 the formula $\mathit{danger}(\mathit{SV},\mathit{POV})$ is a general formula that states that $\mathit{POV}$ poses a danger to $\mathit{SV}$. It is among our key findings that this formula $\mathit{danger}(\mathit{SV},\mathit{POV})$ can be the same in all critical scenarios---scenario-specific features can be expressed in other components of a formula. At the same time, we found that  a good definition of $\mathit{danger}$ is critical to the quality of our formalisation: initially we had a looser condition as $\mathit{danger}$, which matched many traces in which $\mathit{POV}$s are hardly relevant to $\mathit{SV}$ (see also Appendix~\ref{sect: debugger}). 

Our definition of $\mathit{danger}$, described below, is based on the notion of \emph{RSS distance}~\cite{Shalev-Shwartz_Shammah_Shashua_2018}. It is defined as the smallest distance such that, if the lead vehicle brakes at its maximal braking rate, the rear vehicle can avoid a crash by braking at a pre-defined comfortable braking rate. According to~\cite{Shalev-Shwartz_Shammah_Shashua_2018} (see also~\cite{Hasuo22RSSIntro}), the RSS distance is defined by

\begin{align}\label{lon_RSS}\small
    \begin{array}{l}
        \mathit{dRSS}_\mathit{lon}(v_r,v_f):= \max\Bigl(0,v_r \rho + \frac{a_{\mathrm{max}}\rho^2}{2} + \frac{(v_r+a_{\mathrm{max}}\rho)^2}{2b_{\mathrm{min}}} - \frac{v_f^2}{2b_{\mathrm{max}}}\Bigr).
    \end{array}
\end{align}
Here, $v_r,v_f$ are the velocities of the rear and front vehicle respectively, $\rho$ is the reaction time after which the rear vehicle starts braking, $a_{\mathrm{max}}$ is the maximum acceleration of the rear vehicle, $b_{\mathrm{min}}$ is the maximum comfortable braking rate of the rear vehicle, and $b_{\mathrm{max}}$ is the maximum possible braking rate of the front vehicle. 

For multi-lane scenarios, we additionally need a notion of lateral danger. This lateral RSS distance, following~\cite{Shalev-Shwartz_Shammah_Shashua_2018}, is given by
\begin{align}\label{lat_RSS}\small
\begin{aligned}
    & \mathit{dRSS}_{\mathit{lat}}(v_1,v_2) \;:=\; 
 \\&\textstyle\max\Bigl(0,  (v_1 - v_2)\rho + a_{\mathrm{max}}^{\mathit{lat}}\rho^2  + \frac{(v_1 + \rho a_{\mathrm{max}}^{\mathit{lat}})^2 + (v_2 - \rho a_{\mathrm{max}}^{\mathit{lat}})^2}{2b_{\mathrm{min}}^{\mathit{lat}}}\Bigr),
\end{aligned}
\end{align}
where  Vehicle 1 is assumed to be to the left of Vehicle 2, $v_1, v_2$ denote their lateral velocities, $a_{\mathrm{max}}^{\mathit{lat}}$ the maximum lateral acceleration rate the vehicles can apply, and $b_{\mathrm{min}}^{\mathit{lat}}$ their maximum comfortable lateral braking rate. (For simplicity, we ignore the stability factor in~\cite{Shalev-Shwartz_Shammah_Shashua_2018}.)

Using the above functions for RSS distances, we go on to define our formulas for danger. Following the RSS framework, we define a dangerous situation as a violation of the RSS distance between two vehicles.
Our main formula $\mathit{danger}$  is hence defined by 
\begin{align}
    &\mathit{danger}(\mathit{SV},\mathit{POV})\;:=\;\mathcal{G}_{[0,\mathit{minDanger}]} \mathit{rssViolation}(\mathit{SV},\mathit{POV}); \label{eqn: robustdanger}
\end{align} 
it requires the formula $\mathit{rssViolation}(\mathit{SV},\mathit{POV})$ to last at least for the time \textit{minDanger}. The time duration \textit{minDanger} is a parameter that should be suitably chosen; the effect of its choice is investigated in our experiments in \S\ref{sect: experiments}. Note that $\mathit{minDanger}=0$ does \emph{not} imply that there is no danger present; by the semantics of $\mathcal{G}_{[0,0]}$ (see \S\ref{subsect:STL}), it means that $\mathit{danger}(SV,POV)$ is true at time~$T$ if $\mathit{rssViolation}(\mathit{SV},\mathit{POV})$ is true at time $T$; i.e.\ if one chooses $\textit{minDanger}=0$, then $\mathit{danger}(\mathit{SV},\mathit{POV})=\mathit{rssViolation}(\mathit{SV},\mathit{POV})$.

By construction, $\mathit{danger}$ stands for durable danger; the notion of instantaneous danger, used in its definition (\ref{eqn: robustdanger}), is defined as a violation of both the lateral and longitudinal RSS distance by
\begin{align}\label{eqn: rssViolation}\small
\begin{array}{l}
 \mathit{rssViolation}(a,b) \;:=\; \mathit{rssViolation}_\mathit{lon}(a,b) \wedge \mathit{rssViolation}_{\mathit{lat}}(a,b). 
\end{array}
\end{align}
The component formulas therein are defined as follows, using RSS distances from~(\ref{lon_RSS}--\ref{lat_RSS}). 
\begin{align}
  &\small
\begin{array}{l}
 \mathit{rssViolation}_\mathit{lon}(a,b) \;:= 
 \mathit{dangerAhead}(a,b) \vee \mathit{dangerAhead}(b,a),
\end{array}
\\
&\small
\begin{array}{l}
   \mathit{dangerAhead}(a,b) \;:=\; \\
  \quad {s_b - s_a} \ge 0 \wedge {s_b - s_a} \le \mathit{length}(b) + \mathit{dRSS}_\mathit{lon}(v_{lon,a},v_{lon,b});
\end{array}
\\
&\small 
\begin{array}{l}
    \mathit{rssViolation}_{\mathit{lat}}(a,b) \;:=\; \mathit{dangerLeft}(a,b) \vee \mathit{dangerLeft}(b,a), 
\end{array}
\label{danger_lat}
\\
&\small 
\begin{array}{l}
 \mathit{dangerLeft}(a,b) \;:=\; 
 \\
 \quad {d_b - d_a} \ge 0 \wedge {d_b - d_a} \le {\mathit{width}(b) + \mathit{dRSS}_{\mathit{lat}}(v_{lat,b},v_{lat,a})}.
\end{array}
\end{align}
The formulas $\mathit{length}$ and $\mathit{width}$ are from \S\ref{subsect:vehicleConf}. 

Similarly, we introduce 
a durable notion of safety, i.e.\ safety for the duration of $\mathit{minSafe}$. It is called \emph{initSafe} because we use it for the ``safe in the beginning'' requirement  in our template~(\ref{scenarioAsInitSafePlusDisturb}).
\begin{align}
    &\mathit{initSafe}(\mathit{SV},\mathit{POV})\;:=\;\mathcal{G}_{[0,\mathit{minSafe}]} \mathit{instSafe}(\mathit{SV},\mathit{POV}), \label{eqn: initSafe}  \\
&\mathit{instSafe}(a,b) \;:=\; \neg \mathit{rssViolation}(a,b).
\end{align}

When we use robust semantics of STL for \emph{quantitative} notions of truth~\cite{stlrobustsemantics}, we find some modification of the formula $\mathit{rssViolation}$ more useful. See Appendix~\ref{appendix:robustSem}.

\subsection{SV Behaviour}\label{sect: svbehaviour}
For readability we set
\begin{equation*}
    \mathit{laneKeep}(a,L)\;:=\;\mathit{atLane}(a,L),
\end{equation*}    
where $a$ denotes a vehicle, $L$ a lane, and \textit{atLane} is from \S\ref{subsec: vehiclepos}. This formula $\mathit{laneKeep}$ is used as part of $\mathit{behaviourSV}_{i}$ in Table~\ref{table: scenario_formulas}.   

To formalise scenarios in which $\mathit{SV}$ is changing its lane, it is sufficient to specify only the initial or target lane, and that it eventually changes its lane. For a vehicle $a$ and a lane $L$ we thus define 
\begin{align*}
    \mathit{leavingLane}(a,L)&\;:=\; \mathit{atLane}(a,L) \wedge \mathcal{F}(\neg\mathit{atLane}(a,L)), \\
    \mathit{enteringLane}(a,L)&\;:=\; \neg\mathit{atLane}(a,L) \wedge \mathcal{F}(\mathit{atLane}(a,L)).
\end{align*}

\subsection{POV Behaviour}\label{sect: povbehaviour}
To distinguish $\mathit{POV}$'s dangerous cut-in or cut-out maneuvers from regular lane changes (see \S\ref{sect: svbehaviour}), the formulas for the former contain a term $\mathcal{F}\mathit{danger}$ to stress that these behaviours lead to danger. Furthermore, these formulas need the lane currently occupied by the $\mathit{SV}$ as input.
Lastly, we require that $\mathit{POV}$s complete their lane change when (or shortly after) danger occurs, in order to tie the danger to the change of lanes. This is represented by the $ \mathcal{F}_{[0,\mathit{minDanger}]}(\dotsc)$ part of the following formulas. Overall, we obtain
    \begin{align*}\small
        \begin{array}{l}
            \mathit{cutIn}(\mathit{POV},\mathit{SV},L):=\\
            \quad\neg\mathit{sameLane}(\mathit{POV},\mathit{SV},L) 
            \wedge \mathcal{F}\Bigl(\mathit{danger}(\mathit{SV},\mathit{POV}) \\
            \quad \quad \wedge \mathcal{F}_{[0,\mathit{minDanger}]}(\mathit{sameLane}(\mathit{SV},\mathit{POV},L)  \wedge \mathit{aheadOf}(\mathit{SV},\mathit{POV}))\Bigr), \\
            \mathit{cutOut}(\mathit{POV},\mathit{SV},L):=\;\mathit{sameLane}(\mathit{POV},\mathit{SV},L) \\ 
            \quad\wedge \mathcal{F}\Bigl(\mathit{danger}(\mathit{POV},\mathit{SV})\wedge\mathcal{F}_{[0,\mathit{minDanger}]} \neg\mathit{atLane}(\mathit{POV},L)\Bigr). 
        \end{array}
    \end{align*}
Here, \textit{atLane} is from \S\ref{subsec: vehiclepos} and \textit{sameLane} and \textit{aheadOf} are from \S\ref{subsec: relbtwnvehicles}. Note that by the semantics of $\mathcal{F}$ in Table~\ref{table: booleansemantics}, a formula of the form $\mathcal{F}{(\varphi_1\wedge\mathcal{F}_{[0,t]}\varphi_2)}$ is true at time $T$ if 1) there exists a time $T'\ge T$ such that $\varphi_1$ is true at time $T'$ and 2) $\varphi_2$ is true at some time during the interval $[T',T'+t]$.

In ISO 34502~\cite{ISO34502}, it is stated that ``acceleration or deceleration categories actually imply relative velocity differences with respect to the subject vehicle.'' We therefore define our formulas \textit{accel} and \textit{decel} using the \textit{fasterThan} formula from \S\ref{subsec: relbtwnvehicles}. We further add \textit{laneKeep} from \S\ref{sect: svbehaviour} to distinguish this $\mathit{POV}$ behaviour from the dangerous cut-ins and cut-outs defined previously.
\begin{align*}
    &\mathit{accel}(\mathit{POV},\mathit{SV},L):=\; \mathit{fasterThan}(\mathit{SV},\mathit{POV})\wedge \mathit{laneKeep}(\mathit{POV},L), \\
    &\mathit{decel}(\mathit{POV},\mathit{SV},L):=\;\mathit{fasterThan}(\mathit{POV},\mathit{SV})\wedge \mathit{laneKeep}(\mathit{POV},L).
\end{align*}
\subsection{Our Formalisation ISO34502-STL}\label{subsec: ISO34502-STL}
Our set of formulas $\mathit{ISO34502\text{-}STL}:=\{\mathit{scenario}_i\mid i=1,\dotsc,24\}$, describing the traffic disturbance scenarios of Table~\ref{table: scenarios}, is composed of the subformulas defined in \S\ref{subsec: vehiclepos}--\ref{sect: povbehaviour}.  Their construction proceeds in the following steps.

The top-level formula is $\mathit{scenario}_i$; its format is given in (\ref{scenarioAsInitSafePlusDisturb}) (see also the top line of Table~\ref{table: scenario_formulas}). Its first component \textit{initSafe} is common and is defined in (\ref{eqn: initSafe}). For its second component $\mathit{roadSector}_{i}$, we set 
\begin{eqnarray*}\small
    \mathit{roadSector}_i(\mathit{SV},\mathit{POV})
        :=\begin{cases}
        \mathit{mainRoad}(\mathit{SV},\mathit{POV}), &i=1,\dotsc,8, \\
        \mathit{mergeZone}(\mathit{SV},\mathit{POV}), & i=9,\dotsc,16, \\
        \mathit{departZone}(\mathit{SV},\mathit{POV}), &i=17,\dotsc,24, 
    \end{cases} 
\end{eqnarray*} 
with the zone subformulas defined in \S\ref{subsec: vehiclepos}. 

The third component $\mathit{disturb}_i$ is the scenario-specific main part. Its template is given in (\ref{disturbFormat}) (see also the second line of Table~\ref{table: scenario_formulas}). Its components $\mathit{initialCondition}_i$, $\mathit{behaviourSV}_i$, and $\mathit{behaviourPOV}_i$ are shown in Table~\ref{table: scenario_formulas}. Note that the disturbance descriptions are independent of road geometry, thus it suffices to define them for the scenarios $i=1,2, \dotsc, 8$.

Our component subformulas $\mathit{initialCondition}_i$, $\mathit{behaviourSV}_i$, and $\mathit{behaviourPOV}_i$ for each scenario are mostly straightforward conjunctions, following their description in ISO 34502 (see Table~\ref{table: scenarios} and Fig.~\ref{fig:vehiclepositions}). Nevertheless, the following points are worth noting. 
\begin{itemize}
 \item In \textit{Scenario 2}, a three-vehicle cut-out is performed, see \S\ref{subsect: prelim ISO}. Note that this maneuver cannot simply be composed of $\mathit{laneKeep}(\mathit{SV},L)$ and $\mathit{cutOut}(\mathit{POV}_1,\mathit{SV},L)$, since the \textit{cutOut} subformula implies danger between $\mathit{SV}$ and the $\mathit{POV}$ performing the cut-out maneuver (see \S\ref{sect: povbehaviour}), whereas danger in this scenario occurs between $\mathit{SV}$ and $\mathit{POV}_2$.
 \item In nearly all scenarios, the input lane $L$ refers to the lane $\mathit{SV}$ occupies at the beginning of the scenario. The only exception is \textit{Scenario 7}, where we use the lane that $\mathit{SV}$ enters as the input lane. This is due to the fact that in Scenario 7, the $\mathit{SV}$ behaviour is described by the formula $\mathit{enteringLane}$ rather than $\mathit{leavingLane}$, see \S\ref{sect: svbehaviour}.
\end{itemize}

\subsection{Extended Formalisation ISO34502-STL-ext(A)}\label{subsec: ISO34502-STL-ext}
In our experiments (\S\ref{sect: experiments}), we found that natural interpretation of some of the ISO 34502 descriptions potentially overly restrictive. Specifically they concern 1) acceleration/deceleration and 2) relative vehicle positions. Our formalisation \textit{ISO34502-STL} follows such natural (potentially overly restrictive) interpretation; in this section, we suggest its relaxation. We call it \textit{ISO34502-STL-ext}. For experimental comparison of our two formalisations, see \S\ref{subsec: experimentdiscussion}.

The first point of relaxation is about the definition of $\mathit{accel}$ and $\mathit{decel}$. We chose to include vehicle acceleration (resp. deceleration) as an alternative to relative velocity differences and define 
\begin{align*}\small
\begin{array}{rl}
    \mathit{accel}_{\mathit{ext}}(\mathit{POV},\mathit{SV},L)&:=\; (\mathit{fasterThan}(\mathit{SV},\mathit{POV}) \vee \mathit{accelerates}(\mathit{POV})) \\
    &\quad\quad\wedge \mathit{laneKeep}(\mathit{POV},L), \\
    \mathit{decel}_{\mathit{ext}}(\mathit{POV},\mathit{SV},L)&:=\;(\mathit{fasterThan}(\mathit{POV},\mathit{SV})\vee \mathit{decelerates}(\mathit{POV})) \\
    &\quad\quad\wedge \mathit{laneKeep}(\mathit{POV},L), \\
    \mathit{accelerates}(\mathit{POV})&:=\;a_{\mathit{POV}} > 0, \\
    \mathit{decelerates}(\mathit{POV})&:=\;a_{\mathit{POV}} < 0.
\end{array}
\end{align*}

In \textit{ISO34502-STL-extA}, we replace $\mathit{accel}$ and $\mathit{decel}$ by $\mathit{accel}_{\mathit{ext}}$ and $\mathit{decel}_{\mathit{ext}}$. This affects $\mathit{scenario}_i$, $i=3,4,7,8$; all other formulas remain unchanged.

The second point of relaxation is the formula \textit{aheadOf} (cf. \S\ref{subsec: relbtwnvehicles}), in order to (partially) include positions $4$ and $7$ in Fig.~\ref{fig:vehiclepositions}. We define \begin{equation*}
    \mathit{aheadOf}_{\mathit{ext}}(a,b)\;:=\; \mathit{front}(a) < \mathit{front}(b).
\end{equation*}
In \textit{ISO34502-STL-ext}, we further extend \textit{ISO34502-STL-extA} by removing \textit{aheadOf} from the definition of \textit{cutIn} (cf. \S\ref{sect: povbehaviour}) completely (thus redefining dangerous cut-ins to include $\mathit{POV}$ entering $\mathit{SV}$'s lane \emph{behind} $\mathit{SV}$ dangerously), and replacing \textit{aheadOf} by $\mathit{aheadOf}_{\mathit{ext}}$ in all other formulas.

\section{Experimental Evaluation}\label{sect: experiments}
In this section, we seek to quantitatively evaluate our formalisation. In related work on the formalisation of traffic rules in temporal logic, experiments focussed on the percentage of vehicles in a given dataset obeying the formalised traffic rules, see e.g.\ \cite{interstate_althoff, intersection_althoff}. Similarly, we will evaluate our formulas on a set of recorded vehicle traces to gain insight into how often our formulas are true.

We aim to address the following research questions:
\begin{itemize}
    \item[RQ1] Are ISO 34502's original descriptions of traffic disturbance scenarios adequate for \emph{detecting} traffic disturbances?
    \item[RQ2] Are ISO 34502's original descriptions of traffic disturbance scenarios adequate for \emph{classifying} traffic disturbances?
    \item[RQ3] How does our formalisation perform in terms of precision and recall? 
    \item[RQ4] How does the choice of the duration \textit{minDanger} in (\ref{eqn: robustdanger}) affect the recall of our formalisation?
    \item[RQ5] How does our formalisation compare to existing works, in terms of detection of common disturbance scenarios?
\end{itemize} To answer RQ3, we use two sets of ground truths. We will define them later in \S\ref{subsec: experimentdiscussion}.

\subsection{Experiment Setup}\label{subsec: experimentSetup}

We evaluate our formulas for traffic disturbance scenarios on the highD dataset~\cite{highDdataset}, which consists of drone-recorded vehicle traces on main road sections of German highways. The vehicle parameter values needed to define the RSS distance in \S\ref{subsec:danger} are chosen similarly to~\cite{interstate_althoff, formalised_RSS} as $\rho = 0.6$, $a_{\mathrm{max}}=5$, $b_{\mathrm{min}} = 6$, $b_{\mathrm{max}} = 8$, $a_{\mathrm{max,lat}}=1.5$, $b_{\mathrm{min,lat}}=1.5$. We chose $b_{\mathrm{min}}$ such that our RSS distance roughly corresponds to the "halber Tacho" rule-of-thumb used in Germany, stating that on highways, a safe distance (in meters) to the preceding vehicle is given by half the current velocity (in kilometers per hour); e.g.\ at a velocity of $100\unit{\kilo\meter\per\hour}$, we consider $50\unit{\meter}$ a safe distance.

In our evaluation, we first filter the data traces and focus only on those in which there arises a danger in its middle. 
Specifically, we first collect all pairs of cars\footnote{In the highD dataset, a \emph{vehicle} is either a \emph{car} or a \emph{truck}. In our experiments, we focus on cars, for the reason that different physical characteristics of trucks should lead to different scenario descriptions. Elaboration of it is future work.} 
that violate the RSS distance between them eventually. Next, we evaluate the semantics of the formula
\begin{equation}\label{preprocessing formula}
\begin{aligned}
& \mathit{dangerArises}(\mathit{SV},\mathit{POV})\;:=\;    \\
&\quad\mathcal{F} ( \mathit{initSafe}(\mathit{SV},\mathit{POV}) \wedge \mathcal{F}(\mathit{danger}(\mathit{SV},\mathit{POV})) ).
\end{aligned}
\end{equation} over all vehicle pairs $(\mathit{SV}, \mathit{POV})$ found in the previous step. Those traces which do not satisfy $\mathit{dangerArises}(\mathit{SV},\mathit{POV})$ are discarded. Since the implication $\mathit{scenario}_i\Rightarrow \mathit{dangerArises}$ is logically valid, the truth of $\mathit{dangerArises}$ is a necessary condition for any of the formulas in \textit{ISO34502-STL(-ext)} (see \S\ref{subsec: ISO34502-STL}--\ref{subsec: ISO34502-STL-ext}); therefore we are not discarding any traces that are relevant to our formalisation. 
 We then shorten the remaining traces so that they only contain information relevant to our formulas: we discard the part before \textit{initSafe} is true and the part after the last \textit{danger} interval ends. Finally, we discard all data about vehicles other than $\mathit{SV}$ and $\mathit{POV}$.

The result of this filtering of the highD dataset is the set of traces where danger arises, i.e.\ those relevant to the ISO 34502 traffic disturbance scenarios. This trace set is referred to as $\mathit{disturbTraces}$.

The evaluation of our formalisations \textit{ISO34502-STL(-ext)} was conducted by computing the semantics of the formulas $\mathit{scenario}_i$, for $i=1,3,4,\dotsc,8$, over each trace $\sigma\in\mathit{disturbTraces}$. Here, 
\begin{itemize}
 \item we exclude $\mathit{scenario}_2$ since our implementation is limited to two-car scenarios ($\mathit{scenario}_2$ involves three cars and dealing with it is future work); and
 \item we exclude $\mathit{scenario}_9,\dotsc,\mathit{scenario}_{24}$ since the highD dataset only has main road traces.
\end{itemize}

\subsection{Results}\label{subsec: results}
In our filtering of the highD dataset as discussed above, we first identified  99656 pairs $(\mathit{SV}, \mathit{POV})$ 
between which danger according to formula (\ref{eqn: robustdanger}) occurs at some point. Here we used  $\mathit{minDanger}=0$ for the formula $\mathit{danger}$ in~(\ref{eqn: robustdanger}) to be the most inclusive. From these pairs  $(\mathit{SV}, \mathit{POV})$, we created two variations of the set \textit{disturbTraces}, by applying the formula $\mathit{dangerArises}$ in~(\ref{preprocessing formula}) with the following parameter values.
\begin{itemize}
 \item $\textit{disturbTraces}_{1}$: using $\mathit{minSafe}=0.6$ and $\mathit{minDanger}=0$, and
 \item $\textit{disturbTraces}_{2}$: using $\mathit{minSafe}=0.6$ and $\mathit{minDanger}=0.6$.
\end{itemize}

On these trace sets  $\textit{disturbTraces}_{1}, \textit{disturbTraces}_{2}$, we evaluate our sets of formulas \textit{ISO34502-STL} and \textit{ISO34502-STL-ext(A)} (see \S\ref{subsec: ISO34502-STL} and \S\ref{subsec: ISO34502-STL-ext}). In this evaluation, the value of the parameter $\mathit{minDanger}$ in the formulas $\mathit{scenario}_i$ is chosen to match the considered set of traces, that is, $\mathit{minDanger}=0$ for traces in  $\textit{disturbTraces}_{1}$ and $\mathit{minDanger}=0.6$ for traces in  $\textit{disturbTraces}_{2}$. For the lane $L$ in the formulas  $\mathit{scenario}_i$ (\ref{scenarioAsInitSafePlusDisturb}), we use the initial lane of $\mathit{POV}$ for $\mathit{scenario}_7$, and the initial lane of $\mathit{SV}$ for $\mathit{scenario}_i$, $i=1,3,4,5,6,8$.

The results of evaluating the STL formulas $\mathit{scenario}_i$ on the traces in $\mathit{disturbTraces}_j$, $j=1,2$ are collected in Table~\ref{table: experimentresults}. Using our prototype implementation, the evaluation of all formulas $\mathit{scenario}_i$, $i=1,3,\dotsc,8$, over all traces in $\mathit{disturbTraces}_1$ took around $2$ hours on a laptop (Windows 10, Intel Core i7 @ 2.60GHz, 16GB RAM).

\begin{table*}[]
    \caption{Quantitative results of our experiments. \textit{Data trace set T} refers to the set of traces on which our formulas are evaluated. See \S\ref{subsec: experimentSetup}: $\mathit{disturbTraces}_1, \mathit{disturbTraces}_2$ are the sets with $\mathit{minDanger}=0, 0.6$, respectively. 
The number of traces in $T$ is denoted by~$|T|$. In the \textit{spec set} column, the evaluated set of formulas is noted (see \S\ref{subsec: ISO34502-STL} and \S\ref{subsec: ISO34502-STL-ext}). \textit{Matching traces} denotes the total number of traces in $T$ for which at least one of the considered formulas $\mathit{scenario}_i$, $i=1,3,\dotsc,8$ is true. The recall column (see \S\ref{subsec: experimentdiscussion}) refers to recall relative to $|T|$ as ground truth. In each column labelled $s_i$, $i=1,3,\dotsc,8$, we state for how many traces in $T$ the corresponding formula  $\mathit{scenario}_i$ is true.}
    \label{table: experimentresults}
        \begin{tabular}{llll||ll|rrrrrrr}
        \toprule
	 data trace set $T$ & $\mathit{min}$-
& no.\ of   & spec set & matching& recall & 
$\mathit{scenario}_1$ 
& 
$s_3$
&
$s_4$
&
$s_5$
&
$s_6$
& 
$s_7$ 
&
$s_8$
\\
         &$\mathit{Danger}$&traces $|T|$ &  & traces &  & ($s_1$) &  &  &  &  &  &  \\
         \midrule
         $\mathit{disturbTraces}_1$ & $0$& $32398$ & \textit{ISO34502-STL}  & $24038$ & $74.2\%$ & $4091$ & $21941$ & $21941$ & $291$ & $218$ & $3188$ & $924$ \\
          $\mathit{disturbTraces}_1$ & $0$&$32398$ & \textit{ISO34502-STL-extA} & $30288$ & $93.5\%$ & $4091$ & $27652$ & $26008$ & $291$ & $218$ & $4362$ & $969$ \\
          $\mathit{disturbTraces}_1$ &$0$& $32398$ & \textit{ISO34502-STL-ext} & $31139$ &$96.1\%$ & $9364$ & $27652$ & $26008$ & $378$ & $218$ & $4362$ & $969$ \\
          $\mathit{disturbTraces}_2$ & $0.6$&$28881$ & \textit{ISO34502-STL}  & $21564$& $74.7\%$ & $4076$ & $19595$ & $19595$ & $283$ & $215$ & $3171$ & $801$ \\
          $\mathit{disturbTraces}_2$ & $0.6$&$28881$ & \textit{ISO34502-STL-extA} & $27177$ & $94.1\%$ & $4076$ & $24843$ & $23328$ & $283$ & $215$ & $4347$ & $838$ \\
         $\mathit{disturbTraces}_2$ & $0.6$&$28881$ & \textit{ISO34502-STL-ext} & $27963$& $96.8\%$ & $9312$ & $24843$ & $23328$ & $361$ & $215$ & $4347$ & $838$ \\
         \bottomrule
        \end{tabular}
\end{table*}

\subsection{Discussions}
\label{subsec: experimentdiscussion}
In this section, we address our previously stated research questions.

\textit{RQ1: Are ISO 34502's original descriptions of the traffic disturbance scenarios adequate for \emph{detecting} critical scenarios?} 
While the original descriptions are not adequate, our extension (see \S\ref{subsec: ISO34502-STL-ext}) is indeed suitable for the purpose of detecting critical scenarios. By comparing the performance of our formalisations \textit{ISO34502-STL} and \textit{ISO34502-STL-extA} in Table~\ref{table: experimentresults}, it is evident that a significant number of traces are not detected with the strict interpretation of acceleration and deceleration following ISO 34502, suggesting that their definitions should be clarified. With our extension \textit{ISO34502-STL-ext}, we detect over $96\%$ of all traces in \textit{disturbTraces}. This conclusion holds regardless of \textit{minDanger}, see RQ4.

\textit{RQ2: Are ISO 34502's original descriptions of the traffic disturbance scenarios adequate for \emph{classifying} critical scenarios?} 
The descriptions are not adequate for classification purposes. This is to be expected; the ISO 34502 scenario set in Table~\ref{table: scenarios} is \emph{not} designed for the purpose of classification but rather for a comprehensive coverage of all traffic disturbances. As an example, consider $\mathit{SV}$ keeping its lane, while $\mathit{POV}$ performs a dangerous cut-in maneuver while driving at a slightly slower speed than $\mathit{SV}$ throughout the maneuver. This clearly matches Scenario $1$ in Table~\ref{table: scenarios}; however it may also match Scenario $4$ if danger arises shortly after $\mathit{POV}$ has entered $\mathit{SV}$'s lane. Refining the scenario descriptions (and accordingly their formalisation) for the purpose of classification is future work.

\textit{RQ3: How does our formalisation perform in terms of precision and recall?}
We recall the definitions of precision and recall from~\cite{precisionrecall}, namely
$\mathit{precision}={\#\mathit{truePositives}}/{(\#\mathit{truePositives} + \#\mathit{falsePositives}) }$, and $\mathit{recall}= {\#\mathit{truePositives}}/{(\#\mathit{truePositives} + \#\mathit{falseNegatives} )}$.
For our experiments, a \textit{true positive} denotes a disturbance trace for which one of our formulas is true, a \textit{false positive} is a disturbance-free trace for which one of our formulas is true, and a \textit{false negative} denotes a disturbance trace for which none of our formulas are true. Precision and recall for individual scenario-formulas are outside the scope of our experiments, as this would require the dataset to contain labels denoting the type of traffic disturbance scenario.

We have a precision of $100\%$---false positives cannot occur in our formalisation, as our definition of a disturbance trace as a trace for which the formula \textit{dangerArises} (see \S\ref{subsec: experimentSetup}) is eventually true has been exactly implemented into all our formulas.

The recall of \textit{ISO34502-STL}, and particularly of \textit{ISO34502-STL-ext}, relative to the sets $\mathit{disturbTraces}_i$, $i=1,2$ is very high, see Table~\ref{table: experimentresults}. This notion of recall is exactly what we discussed in RQ1.

We note that the use of \textit{initSafe} (see \S\ref{subsec:danger}) is important. Indeed, if we omit the filtering using \textit{dangerArises} (\ref{preprocessing formula}) and merely filter using $\mathit{danger}$ (\ref{eqn: robustdanger}), our recall is quite low (around $25\%$) due to many false negatives. This is natural; due to the spatial limitation of the highD dataset, many traces do not fulfill \textit{initSafe}, i.e.\ danger is either present from the beginning of the trace or the observed interval of safety leading up to danger is shorter than required by \textit{initSafe}. This spatial limitation can be considered a perception disturbance (see \S\ref{subsect: prelim ISO}), which is outside the scope of our formalisation.

\textit{RQ4: How does the choice of \textit{minDanger} in (\ref{eqn: robustdanger}) affect recall of our formalisation?}

The choice of \textit{minDanger} in (\ref{eqn: robustdanger}) does not have a significant impact on recall, see Table~\ref{table: experimentresults}. For this reason, we restrict our discussion of the other research questions to $\mathit{minDanger}=0$. 

One implication of the above  is that our formalisation---based on the RSS-based notion of danger---is \emph{stable}, with respect to a parameter value which is not easy to choose (namely $\mathit{minDanger}$).

\textit{RQ5: How does our formalisation compare to existing works, in terms of detection of common disturbance scenarios?
}

\begin{table}[]
    \caption{Absolute number of the cut-in and deceleration scenarios detected in the highD dataset by \cite{highDMonitoring,highDsakuracomparison} and our formalisation (see \S\ref{subsec: ISO34502-STL}--\ref{subsec: ISO34502-STL-ext}). Definitions and danger metrics of~\cite{highDMonitoring} and~\cite{highDsakuracomparison} are discussed in the discussion of RQ5 in \S\ref{subsec: experimentdiscussion}. Our danger metric \emph{RSS distance} is defined in \S\ref{subsec:danger}. The values in the \textit{ISO34502-STL(-ext)} columns are from Table~\ref{table: experimentresults}. For details see the discussion of RQ5 in \S\ref{subsec: experimentdiscussion}.
}\label{table: comparisonhighD}
    \begin{tabular}{lrrrr}
        \toprule
    & \cite{highDMonitoring} & \cite{highDMonitoring}  & \cite{highDsakuracomparison}  & ISO34502 -STL-ext \\
    \midrule
    danger metric & TTC & THW  & none  & RSS dist.  \\
    cut-in       & $144$  & $7219$ & $1017$ & $9364$  \\
    deceleration & $22$  & $112$  & $26846$ & $26008$  \\
    \bottomrule
    \end{tabular}
\end{table}

Here we compare the content of our formalisation with some existing works. We focus on the cut-in and deceleration scenarios,
as these two scenarios have been studied many times before. Our comparison is with \cite{highDMonitoring} (with two danger metrics $\mathrm{TTC}$ and $\mathrm{THW}$) and~\cite{highDsakuracomparison}, as these works 
evaluate their formalisation on the highD dataset. The comparison is in Table~\ref{table: comparisonhighD}; here are some details.

In~\cite{highDMonitoring}, a cut-in is defined as a lane change in which $\mathit{POV}$ 
 enters the lane of $\mathit{SV}$ in front of $\mathit{SV}$. The notion of deceleration is defined in~\cite{highDMonitoring} as hard braking (above a parameter value they choose) by the $\mathit{POV}$ ahead of $\mathit{SV}$ in the same lane. As danger metrics, \textit{time to collision} ($\mathrm{TTC} <4\unit{\second}$) and \textit{time headway} ($\mathrm{THW} <2\unit{\second}$) are used.

In~\cite{highDsakuracomparison}, a cut-in is defined as in~\cite{highDMonitoring}, with the additional restrictions that 1) the lateral velocity of $\mathit{POV}$ must remain nonzero throughout the maneuver, and 2) a maximum longitudinal distance between $\mathit{SV}$ and $\mathit{POV}$ is introduced. A deceleration scenario is defined as ``reduction
of the headway distance between vehicles [...] caused by
the deceleration of a preceding vehicle and not by the
acceleration of the subject vehicle'', where $\mathit{POV}$ is required to brake continuously throughout the scenario and to remain in its lane. In~\cite{highDsakuracomparison}, matching traces are first detected without any danger metric; later the danger metrics $\mathrm{TTC}$ and $\mathrm{THW}$ are calculated for the detected traces. We restrict our discussion to the number of traces that were detected without any danger metric, as their danger metric analysis lacks absolute numbers.

For our formalisation, we use ISO34502-STL-ext (\S\ref{subsec: ISO34502-STL-ext}), based on our discussion in RQ1.
We identified cut-in as Scenario 1 and deceleration as Scenario 4: although Scenarios 5 and 8 also contain cut-in and deceleration components, respectively, Scenarios 1 and~4 correspond more closely to the definitions used in~\cite{highDMonitoring,highDsakuracomparison}. The numbers for ISO34502-STL-ext in Table~\ref{table: comparisonhighD} are hence the numbers for Scenario~1 (cut-in) and Scenario~4 (deceleration) from the third row of Table~\ref{table: experimentresults}, i.e.\ for $\mathit{minDanger}=0$ and ISO34502-STL-ext.

In Table~\ref{table: comparisonhighD}, the numbers for different formalisations for both the cut-in and deceleration scenarios are quite different. For each scenario, the number for our formalisation is (close to) the greatest, and we may wonder if the number is too large. We argue that this is not the case: 
\begin{itemize}
 \item the trace set $\mathit{disturbTraces}_{1}$ is a naturally defined one by the formula $\mathit{dangerArises}$ in (\ref{preprocessing formula}); 
 \item the whole set ISO34502-STL-ext of formulas, achieving $96.1\%$ recall 
for $\mathit{disturbTraces}_{1}$ (see RQ3), is therefore a natural one (note that precision is $100\%$ by definition); and
 \item the formulas $\mathit{scenario}_1$ and $\mathit{scenario}_4$, defined by imposing natural constraints (Table~\ref{table: scenario_formulas}), are natural ones, too.
\end{itemize}

We note that
a similar 
number
of cut-ins as in our experiments 
is detected in~\cite{highDMonitoring} ($7219$). This is natural, since their danger metric $\mathrm{THW} <2\unit{\second}$ is very similar to the ``halber Tacho'' rule we used to calibrate our parameter values for the RSS distance (see \S\ref{subsec: experimentSetup}) and their definition of a cut-in is similar to ISO 34502. On the other hand, \cite{highDsakuracomparison} detects significantly less cut-ins ($1017$) even without a danger metric; this may be due to the restriction on the longitudinal distance between the two vehicles.

For deceleration scenarios,~\cite{highDsakuracomparison} detects a very similar number ($26846$) as our formalisations. 
Note that in~\cite{highDsakuracomparison}, no danger metric is used, implying that deceleration on highways leads to a violation of the RSS distance (see \S\ref{subsec:danger}) in nearly all cases. On the other hand, the formalisation in~\cite{highDMonitoring} is much stricter and hence matches significantly fewer traces ($<150$).

Another point in the comparison
is that \emph{parameter-tuning is easy} in our formalisation. Our parameter-tuning is largely reduced to that for RSS in general: 
\begin{itemize}
 \item besides the RSS parameters (e.g.\ response time $\rho$ and maximum acceleration/braking rates, see \S\ref{subsec:danger}), our formalisation only has $\mathit{minDanger}$ and $\mathit{minSafe}$ as parameters, and
 \item as seen in RQ4, the value of $\mathit{minDanger}$ is hardly relevant. 
\end{itemize}
Tuning the RSS parameters is a more general problem---those parameters are important  beyond specific disturbance scenarios---and methods of doing so are an active field of research~\cite{KonigshofOSS22}. Such ease of parameter tuning is not the case, e.g.\ in~\cite{highDMonitoring}, where ``hard braking'' in a deceleration scenario must be manually defined.

\balance
\section{Conclusions and Future Work}\label{sect: conclusion}
In this paper, we presented a temporal logic formalisation of traffic disturbance scenarios concerning general vehicles on highways as specified in ISO Standard 34502. Based on this important class of scenarios, further critical scenarios can be formalised through a modular approach. Our formalisation relies on the RSS framework, allowing for easy parameter-tuning. 
Experimentally, we were able to show that a small extension of the ISO standard allows for a very high rate of disturbance trace detection.

Future work includes the formalisation of further critical scenarios in ISO 34502 and beyond, such as scenarios involving motorcycles, vehicle control disturbances, and intersection scenarios.

Our current formalisation needs an initial lane as input, since STL does not support variable binding and hence does not allow us to refer to the lane initially occupies otherwise. As future work, an implementation of our formulas to an extension of STL with temporal freezing (e.g.\ \emph{STL*}, see~\cite{stlstar}) should be considered.



\clearpage

\nobalance

\appendix

\section{Road Network}\label{appendix:roadNetwork}
We now give a rigorous description of our modelling of the road network using lanelets. This description is based on~\cite{interstate_althoff}.

A lanelet $l$ has an associated \textit{zone} $$\mathit{zone}(l)\in\{\mathit{mainRoad}, \mathit{mergeZone}, \mathit{ departZone}\},$$ describing the road sector it belongs to, and an \textit{attribute} $$\mathit{attr}(l)\in\{\mathit{main}, \mathit{merge}, \mathit{departure}\},$$ describing the function of its lane (to be defined later in this section).
Note that it is possible to have $\mathit{attr}(l)=\mathit{main}$ but $\mathit{zone}(l)= \mathit{ departZone}$, e.g.\ for a lanelet that is part of a main lane 
and located near the beginning of a departure lane, see Fig.~\ref{fig:lanelets}.

We let $\mathit{leftBound}(l)$ and $\mathit{rightBound}(l)$ denote the boundaries of a lanelet $l$, defined as sets of points. Additionally, we define the start- and endpoints of the boundary polylines as $\mathit{start}(l)$ and $\mathit{end}(l)$. 
We say that a lanelet $l_1$ is \emph{left-adjacent} to another lanelet $l_2$ if it holds that $\mathit{leftBound}(l_2)=\mathit{rightBound}(l_1)$. Accordingly, $l_2$ is \emph{right-adjacent} to $l_1$ if $l_1$ is left-adjacent to $l_2$. Furthermore, we say that $l_1$ is the \emph{predecessor} of $l_2$ if $\mathit{start}(l_2)=\mathit{end}(l_1)$ and $\mathit{attr}(l_1)=\mathit{attr}(l_2)$. Analogously, $l_1$ is the \emph{successor} of $l_2$ if $l_2$ is the predecessor of $l_1$. For rigorous definitions, see~\cite{interstate_althoff}.

A \textit{lane} $L$ is defined as the set of all lanelets preceding and succeeding each other and is defined by recursion; if $\mathit{pred}(l)$ or $\mathit{succ}(l)$ does not exist for a lanelet $l$, we use $\mathit{pred}(l):=\emptyset$ (resp.\ $\mathit{succ}(l):=\emptyset$).
\begin{eqnarray*}
    \mathit{lane}(l)\;:=\;
        \{l\}\cup\mathit{lane}(\mathit{pred}(l))\cup\mathit{lane}(\mathit{succ}(l)).
\end{eqnarray*} We assume that predecessors and successors are unique, hence a single lane cannot lead in two different directions.

We say that lane $L_2$ is to the left (resp. right) of lane $L_1$ if
\begin{align*}
    \mathit{adj}_l(L_1,L_2) &\;:=\; \textstyle\bigvee_{l_1\in L_1}\bigvee_{l_2\in L_2} l_2 \text{ left-adjacent to } l_1, \\
    \mathit{adj}_r(L_1,L_2) &\;:=\; \textstyle\bigvee_{l_1\in L_1}\bigvee_{l_2\in L_2} l_2 \text{ right-adjacent to } l_1,
\end{align*}
and define the set of all lanes adjacent to lane $L$ as \begin{align*}
    \begin{array}{c}
    \mathit{adjLanes}(L)\;:=\;\{ L' \text{ lane} \mid (L'\ne L) \wedge ( \mathit{adj}_l(L,L')\vee  \mathit{adj}_r(L,L') ) \}.
\end{array}
\end{align*}

\section{STL Debugger}\label{sect: debugger}
\begin{figure}[tbp]
    \includegraphics[width=0.45\textwidth]{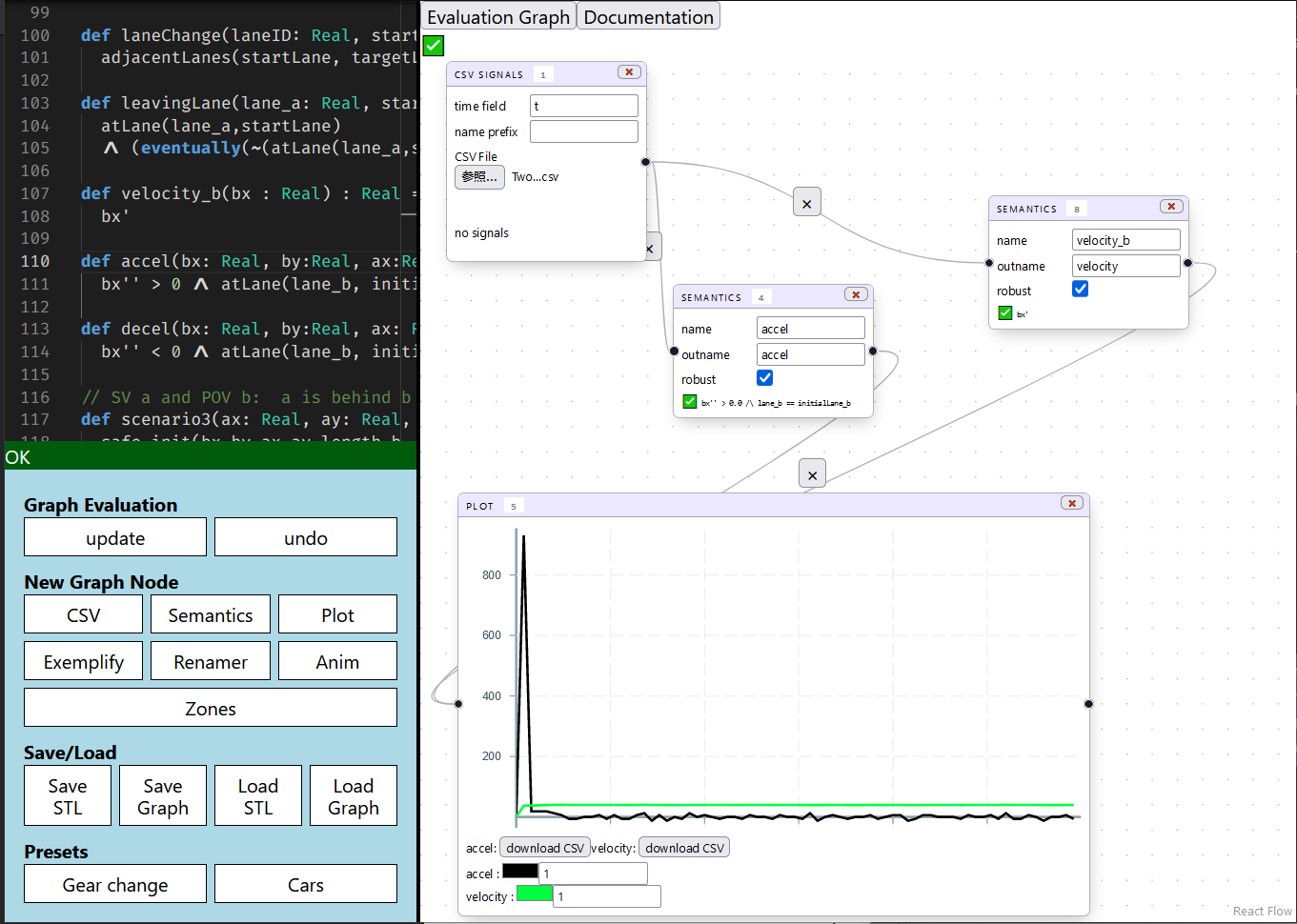}
    \caption{STL Debugger, a screenshot. STL formulas are written in the text pane (left). Functionalities such as semantics computation and exemplifications are realised in the GUI on the right, where different nodes are connected with wires (they can be dragged for rearrangement) and results can be plotted or even animated. 
}
    \label{fig:debugger}
\end{figure}

In the course of devising the STL formulas in \textit{ISO34502-STL} (Table~\ref{table: scenario_formulas}), we benefited from a  tool called \emph{STL Debugger}. It is a browser-based interactive tool for ``debugging'' STL formulas, developed by some of the authors and their colleagues. Its screenshot is in Fig.~\ref{fig:debugger}. 

STL Debugger is developed for  the following purpose. STL~\cite{stl_definition}, much like other temporal logics, has advantages in its expressivity, succinctness, and readability. STL is expected to serve as a basis of many monitoring, testing and verification frameworks, providing them a uniform interface for users to express the events (properties, requirements, etc.) of their interest. Those frameworks will automate common analysis tasks in the cyber-physical systems domain, and help a number of industry practitioners. 

However, STL is a  formalism that is new to many such practitioners. Therefore it can still be challenging---at least in the beginning---for them to come up with \emph{right} STL formulas, those which adequately express their intent. In STL Debugger,  a user can \emph{write} an STL formula, \emph{inspect} its meaning, often \emph{discover the gap} between its meaning and their (informal) intent, and \emph{revise} the formula. Iterating this loop will eventually lead to a right STL formula.

In this work we used STL Debugger to refine trial STL formalisations to eventually reach the formulas in Table~\ref{table: scenario_formulas}. In this way, we found that STL Debugger is useful not only for STL beginners but also for people with logic backgrounds (such as the authors). 

STL Debugger offers two main functionalities. The first is \emph{computing semantics}: given an STL formula $\varphi$ and a signal $\sigma$, it can compute the Boolean truth (whether $\sigma\models\varphi$) as well as the (quantitative) robust semantics (cf.\ \S\ref{subsect:STL}). If the truth value is different from what the user expects, then it implies a gap between the formula and the user's intent, and thus that the formula must be modified. 

STL Debugger's second main functionality is \emph{exemplification}. Given an STL formula $\varphi$, the tool synthesises a signal $\sigma$ that satisfies $\varphi$ (i.e.\ $\sigma\models\varphi$). Often the synthesised signal $\sigma$ is---according to the user's intuition---one that should \emph{not} satisfy the specification. In this case, again, the formula $\varphi$ must be modified. (A similar functionality is  in STLInspector from~\cite{RoehmHM17}.)

These functionalities of STL Debugger have been useful in the current work. For one, our initial trials had a much looser definition of the formula $\mathit{danger}$ (\S\ref{subsec:danger}). The exemplification functionality of STL Debugger quickly pointed to this problem by synthesising a signal in which two vehicles are too far away to be relevant to each other. This eventually led to our current RSS-based definition of  $\mathit{danger}$. For another, the combination of textual and graphical interfaces in STL Debugger allowed us to easily inspect not only the semantics of a formula as a whole but also the semantics of its constituent parts (\emph{subformulas}). This feature guided us to modular development of STL formalisation, leading to the common templates shown in~(\ref{scenarioAsInitSafePlusDisturb}--\ref{disturbFormat}). 

The flexibility of the GUI also enabled us to compare the semantics of the formulas $\varphi_{\mathit{before}}$ and  $\varphi_{\mathit{after}}$ before and after modification, speeding up our search for right formulas. The exemplification functionality was used for highlighting the difference between  $\varphi_{\mathit{before}}$ and  $\varphi_{\mathit{after}}$---exemplifying $\varphi_{\mathit{before}}\land\lnot\varphi_{\mathit{after}}$ (or its negation) synthesises a signal that resides in their semantical difference.

\section{Robust Semantics}\label{appendix:robustSem}
Besides the standard setting of (qualitative) Boolean truth values, a quantitative \emph{robust semantics} is also widely used. In robust semantics, truth values are real-valued numbers, where 1) a positive value stands for \textit{true} and a negative value for \textit{false}; and 2) their absolute values stand for the \emph{margins} of being so. These semantics are defined inductively in the same manner as the Boolean semantics; 
see e.g.~\cite{stlrobustsemantics}.
Robust semantics have the advantage of not only telling us \textit{if} a formula is true, but also including information about how robustly a formula is true.

\begin{remark}
It is natural to use robust semantics (cf.\ \S\ref{subsect:STL}) for the  formula
$\mathit{rssViolation}$ (\ref{eqn: rssViolation})
to measure the degree of danger. The above definition of the formula, however, is not suited for the purpose. It violates a natural \emph{monotonicity} property---between the robust semantics and the inter-vehicle distance---due to a specific way of interpreting $\land$ in robust semantics. (The interpretation of Boolean connectives such as $\land$ is a general problem in robust semantics. See e.g.~\cite{DBLP:conf/cav/ZhangHA19}.) 

A possible fix is to use the following formulas. They are equivalent to the above definition in terms of signs (thus Boolean truth), but are better behaved with respect to quantitative robust semantics as defined in \S\ref{subsect:STL}. Here we set $\tfrac{1}{0}:=\infty$.
\begin{multline*}
    \mathit{dangerAhead}_{rs}(a,b) \;:=\; \\ \textstyle\frac{1}{s_b - s_a} \ge 0 \wedge \textstyle\frac{1}{s_b - s_a} \ge \frac{1}{\mathit{length}(b) + \mathit{dRSS}_\mathit{lon}(v_{lon,a},v_{lon,b})},
\end{multline*}
\begin{multline*}
    \mathit{dangerLeft}_{rs}(a,b) \;:=\; \\ \textstyle\frac{1}{d_b - d_a} \ge 0 \wedge \frac{1}{d_b - d_a} \ge \frac{1}{\mathit{width}(b) + \mathit{dRSS}_{\mathit{lat}}(v_{lat,b},v_{lat,a})}.
\end{multline*}
\end{remark}
\end{document}